\documentclass[letterpaper, 10 pt, conference]{ieeeconf}  

\IEEEoverridecommandlockouts                              

\usepackage{amsmath} 

\usepackage{subfigure}
\usepackage{caption}
\usepackage{makecell}
\usepackage{amsmath} 
\usepackage{multirow}
\usepackage{graphicx}
\usepackage{xcolor}
\usepackage{amssymb}

\title{\LARGE \bf
DA$^{\textbf{2}}$-Net : Diverse \& Adaptive Attention Convolutional Neural Network
}

\author{Abenezer Girma, Abdollah Homaifar$^*$ , M Nabil Mahmoud, Xuyang Yan and Mrinmoy Sarkar 
\thanks{ All the authors are with Faculty of Electrical and Computer Engineering,
        North Caroliina A\&T State University, 1601 E Market St, Greensboro, NC 27411
        {\tt\small Email: aggirma@aggies.ncat.edu,  homaifar@ncat.edu, mnmahmoud@ncat.edu, xyan@aggies.ncat.edu and msarkar@aggies.ncat.edu}}%
\thanks{*Corresponding Author: A. Homaifar, Telephone: (336) 2853271.}
}
\begin{document}
\maketitle
\thispagestyle{empty}
\pagestyle{empty}
\begin{abstract}
Standard Convolutional Neural Network (CNN) designs rarely focus on the importance of explicitly capturing diverse features to enhance the network's performance. Instead, most existing methods follow an indirect approach of increasing or tuning the networks' depth and width, which in many cases significantly increase the computational cost. Inspired by biological visual system, we proposes a Diverse and Adaptive Attention Convolutional Network (DA$^{2}$-Net), which enables any feed-forward CNNs to explicitly capture diverse features and adaptively select and emphasize the most informative features to efficiently boost the network's performance. DA$^{2}$-Net incurs negligible computational overhead and it is designed to be easily integrated with any CNN architecture. We extensively evaluated DA$^{2}$-Net on benchmark datasets, including CIFAR100, SVHN, and ImageNet, with various CNN architectures. The experimental results show DA$^{2}$-Net provides a significant performance improvement with very minimal computational overhead.
\end{abstract}
\section{Introduction}
{\color{green!55!blue} }
A biological visual system has been the primary inspiration for many modern CNN architecture construction. Recent neurological studies \cite{mohan2019diversity,vidyasagar2015origins}, have shown the cortical neurons enables the visual cortex system to explicitly capture and adaptively process diverse resolution of features. The term "diverse" in this work focus specifically on different resolution and size of image features. Additionally, recent studies \cite{szegedy2017inception,wang2018learning} have shown the importance of capturing diverse features in increasing CNN's representational power. However, many current standard CNN architectures use implicit standard feature extractor approach which is not effective in explicitly capturing diverse features. As a result enhancing CNNs performance following standard approach usually results in high computational cost increase \cite{howard2017mobilenets,he2016deep,ren2015faster,huang2017densely,simonyan2014very,szegedy2015going,zagoruyko2016wide}. In this paper we aim to overcome this limitations using attention mechanisms approach to explicitly capture diverse features and efficiently boost CNN's performance.

\begin{figure*}[t]
    \centering
    \includegraphics[width=\textwidth]{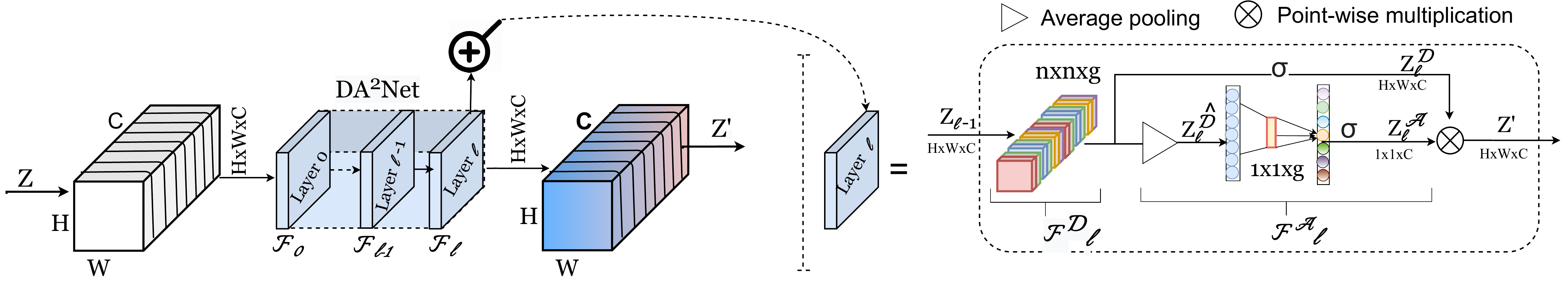}
    \caption{DA$^{2}$-Net takes an intermediate-feature map $Z$ and sequentially conduct diverse feature extraction $\mathcal{F}^{\mathcal{D}}$ and adaptive feature selection $\mathcal{F}^{\mathcal{A}}$. The right side of the picture shows how DA$^{2}$-Net integrates with the backbone architecture, and the left side shows the detail of DA$^{2}$-Net's structural design. }
    \label{fig:DAM_plug}
\end{figure*}

Recently, the use of attention in CNN has attracted many researchers' interest. For instance, Gather-Excite (GE) \cite{hu2018gather} utilize attention via a gather operation to aggregate feature map response and excite operation to combine the pooled information with the feature-map. Squeeze-and-Excitation (SE) \cite{hu2018squeeze} uses an attention mechanism to re-scale feature-map in intermediate layers. However, both SE and GE only focus on contextual information while ignoring the importance of diversity of the feature response. BAM\cite{park2018bam} uses max and average pooling and applies standard convolution to capture general feature descriptors. On the other hand, CBAM\cite{woo2018cbam} omits to use max and average pooling as in BAM\cite{park2018bam}, instead directly applies standard convolution with dilation. However, both BAM and CBAM intend to learn general spatial contextual information, failing to capture and select diverse features. In other words, the standard convolution \cite{goodfellow2016deep} used in BAM and CBAM is not designed to capture different patterns in the image with various resolutions. Most recently studies show success in the use transformer types of attention for computer vision tasks, however such approach requires high computation cost. In \cite{dosovitskiy2020image} transformer requires 210x and 25x number of parameters than MobileNet \cite{howard2017mobilenets} and ResNet-50 \cite{he2016deep} architectures respectively. As a result current transformer based approach are not optimized for efficient performance boost.  Another study in \cite{komodakis2017paying} uses an attention mechanism to transfer knowledge from teacher CNN to student network. However, its' integration with CNN requires significant modification of the backbone architecture.

Some standalone architectures attempt to address the feature diversity problem partially. For instance, SKNet \cite{li2019selective} proposes a standalone CNN architecture with multiple branches and  kernel sizes aming to change the neurons' receptive field in the network. Nevertheless, SKNet does not intend to capture diverse features; instead, it aims to flexibly adjust the neurons' receptive field in CNN to increase effective receptive field area. Besides, the implementation of multiple branches used in SKNet design significantly increases the model's time complexity, resulting in high latency during inference. Also, unlike attention mechanism, SKNet is a standalone architecture which is not intended to be integrated with other CNN architectures. 
  
In this paper, we propose a Diverse and Adaptive Attention for Convolutional Networks (DA$^{2}$-Net), aiming to capture and adaptively select diverse features to efficiently boost CNNs performance. DA$^{2}$-Net is designed to be computationally light-weight and easy to integrate with any CNN architectures. As shown in Fig \ref{fig:DAM_plug}, DA$^{2}$-Net uses multiple sizes filters connected sequentially to effectively capture diverse features and refine it layer by layer. DA$^{2}$-Net then learns the non-linear inter-feature-maps relationship among the captured features to adaptively selects the most informative features. We aim to learn feature-maps interaction at the local level among $\alpha$ neighboring feature-maps out of $n$ available feature-maps. Our analysis shows that this strategy can provide better efficiency and effectiveness.  Further, DA$^{2}$-Net is designed to have the filters placed sequentially in an increasing order based on thier size to leverage the advantage of wider filters in maximizing the Effective Receptive Field (ERF) \cite{luo2017understanding}. In both diverse feature extraction and adaptive selection transformation, we avoid to use point-wise convolution for dimensionality reduction, which weakens the direct relations among feature-maps. Instead, we use a group-wise convolution to achieve better efficiency and performance improvement. 

 To verify the effectiveness of the proposed approach we conducted an extensive experimental analysis by integrarting DA$^{2}$-Net with various CNN architectures and using benchmark datasets; including CIFAR-$100$ \cite{krizhevsky2009learning}, SVHN \cite{netzer2011reading} and ImageNet-$1$K\cite{krizhevsky2012imagenet}. As shown in Experiment Section \ref{sec:experiment}, DA$^{2}$-Net improves the performance of the baseline models and outperforms other attention mechanism with negligible increase in computational overhead. In summary, the main contributions of this paper are listed as follows:
\begin{itemize}
    \item  We propose an effective attention mechanism, DA$^{2}$-Net, to efficiently boost CNN networks' performance. Unlike current CNN attention mechanisms \cite{park2018bam,woo2018cbam,hu2018squeeze,hu2018gather}, DA$^{2}$-Net can capture and select diverse informative features to efficiently boost CNN performance. 
    
    \item DA$^{2}$-Net avoids dimensionality reduction that is used in \cite{park2018bam,woo2018cbam,hu2018squeeze,hu2018gather}, which reduces the linear relation among the feature-map; instead, it uses group-wise convolution to achieve better efficiency and accuracy. 
  
    \item We integrated DA$^{2}$-Net with existing CNN architectures such as \cite{he2016deep,huang2017densely,zagoruyko2016wide,xie2017aggregated, li2019selective} which significantly improves the performance with negligible computational overhead increase. DA$^{2}$-Net also outperforms other state-of-the-art   attention mechanisms \cite{park2018bam,woo2018cbam,hu2018gather}.  

\end{itemize}
The remainder of this paper is organized as follows: Section \ref{sec:methodolody} presents details about the proposed method. Section \ref{sec:experiment} presents the experimental results and the conclusion is presented in Section \ref{sec:conclusion}.
\section{Methodology : DA$^{2}$-Net } \label{sec:methodolody}
{\color{green!55!blue} }
In this section, we discuss the details of the proposed approach.

\subsection{Overview of  DA$^{2}$-Net}
As shown in Fig \ref{fig:DAM_plug}, DA$^{2}$-Net takes an intermediate feature-map $Z_{i} \in \mathbb{R}^{H\times W\times C}$ from layer $i$ of backbone architecture and conducts an attention transformation $Z\textquotesingle_{i} = \mathcal{F}(Z_{i})$ using the following steps. \textbf{First}, it conduct a diverse feature extraction $\mathcal{F}^{\mathcal{D}}$ to effectively capture different resolutions of features from the intermediate feature-map. \textbf{Second}, it applies adaptive selection $\mathcal{F}^{\mathcal{A}}$ to learn inter-feature-map relationship among the captured features and generate weighing values. \textbf{Third}, it uses the generated weighting values to adaptively select/emphasis the most informative feature-maps before passing the feature-maps to the next layer. This process takes place at each layer of DA$^{2}$-Net to effectively capture and refine diverse resolution of features. 

Assuming DA$^{2}$-Net has $\ell$ layers, the overall attention transformation can be represented as:

\begin{equation}
 \begin{split}
  Z\textquotesingle &= \mathcal{F}_{\ell}\odot...\odot\mathcal{F}_{2}\odot\mathcal{F}_{1}(Z_{i}),\\
    Z\textquotesingle &=
  \bigodot_{j=1...\ell}\mathcal{F}_{j}(Z_{i}), 
    \end{split}
    \label{eq:atten_trans}
\end{equation}

where $\mathcal{F}_{1},\mathcal{F}_{1},...,\mathcal{F}_{\ell}$ denotes the layer by layer transformation in DA$^{2}$-Net. The symbol $\odot$ represents layers connection as $\mathcal{F}_{2} \odot \mathcal{F}_{1} = \mathcal{F}_{2}(\mathcal{F}_{1})$, and $Z\textquotesingle \in \mathbb{R}^{H\times W\times C}$ denotes the final output of DA$^{2}$-Net. Equation \ref{eq:atten_trans} can be expressed in more detail using DA$^{2}$-Net sub-components; diverse feature extraction $\mathcal{F}^{\mathcal{D}}$ and adaptive feature selection $\mathcal{F}^{\mathcal{A}}$ as:
\begin{equation}
\begin{gathered}
  Z\textquotesingle = \mathcal{F}^{\mathcal{A}}_{\ell}(\mathcal{F}^{\mathcal{D}}_{\ell} (Z_{\ell-1} ))  \odot...\odot \mathcal{F}^{\mathcal{A}}_{2}(\mathcal{F}^{\mathcal{D}}_{2} (Z_{1} ))  \odot \mathcal{F}^{\mathcal{A}}_{1}(\mathcal{F}^{\mathcal{D}}_{1} (Z_{i} )) \\[1ex]
      Z\textquotesingle =
  \bigodot_{j=1...\ell}\mathcal{F}^{\mathcal{A}}_{j}(\mathcal{F}^{\mathcal{D}}_{j} (Z_{i-1} )) 
    \end{gathered}
    \label{eq:atten_trans2}
\end{equation}

The details of diverse feature extraction $\mathcal{F}^{D}$ and adaptive selection $\mathcal{F}^{A}$ are discussed in the following subsections.

\subsection{Diverse feature extraction}
{\color{green!55!blue} }
Capturing diverse feature in CNN is important to increase the discriminative power of the model \cite{szegedy2017inception,wang2018learning}. Increasing the depth or the width of the network indirectly enable CNN to increase its ability to capture more diverse features and improve the performance, however this approach results a high computational cost increase \cite{szegedy2015going,he2016deep}. Prior work has shown that using multiple size of filters can be effective at capturing various resolution of information for visual tasks \cite{szegedy2017inception,tan2019mixnet}. However, majority of widely used CNN architectures \cite{howard2017mobilenets,he2016deep,ren2015faster,huang2017densely,simonyan2014very,szegedy2015going,zagoruyko2016wide}  do not take advantage of this finding. They use a uniform size of filter in their architectural design. Noticing the gap and considering the effectiveness of multi-size filter approach, we aim to design multiple size of filters as an attention mechanism to capture various resolution of features and efficiently boost any feed-forward CNN architecture performance.

Our approach considers applying diverse feature extraction for each feature-map or a few neighboring feature-maps separately. Unlike typical feature-extraction techniques, this helps to separately examine each feature-map or neighboring feature maps without mixing inter-feature-map information. To achieve these, we use a group-wise convolution with different filter sizes, where the group $G$ can range from individual feature-map $G=1$ to a sub-set of feature-maps $G=c/g$, where $c$ is number of feature maps and $g$ is the grouping ratio or the number of grouped adjacent feature-maps. Note that fully or partially preventing inter-feature-map information mix is essential for the subsequent adaptive selection phase, where we learn inter-feature-map relationships separately. 

The group convolution in each layer of DA$^{2}$-Net uses different size of the filters $n\times n$, where $n=2x +1$ and $x$ is integer ranging in $1\leq x \leq 4$. This enables to incorporate up-to four different sizes of filters in DA$^{2}$-Net. Note that, as the intermediate feature map $Z$ passes from layer to layer in DA$^{2}$-Net, while getting filtered and refined by different filter sizes at each layer. Thus, enabling the network to effectively capture diverse features.
We set the value of $x$ below four to control the filter size from becoming to large and be equal to the size of feature map, which makes the convolution operation to become a simple multi-layer perceptron. Hence, we limit the filter sizes in-between three to nine by setting $x \leq 4$. Additionally, we place the filters in an increasing order because large filters is capable of increasing the Effective Receptive Field (ERF) of the network \cite{luo2017understanding}. Bigger ERF increases the network's representational power, especially for applications such as image segmentation \cite{noh2015learning}.

We implement the above strategy using a grouped depthwise-separable convolution $f^{n\times n}_{G}$ \cite{mamalet2012simplifying}, with multiple size of filters ($n\times n$) and $G$  number of groups. The separable convolution is then followed by batch normalization (BN)  \cite{ioffe2015batch} and sigmoid activation function ($\sigma$) to finally generate the partially transformed feature map  $ Z_{\ell}^{\mathcal{D}} \in \mathbb{R}^{H\times W\times C}$, which can be represented as:
\begin{equation}
\begin{gathered}
 \footnotesize
    Z_{\ell}^{\mathcal{D}} =  \sigma(BN(f^{n\times n}_{G}( Z_{\ell-1 }) ) ) 
    \label{eq:spatial_atten}
\end{gathered}
\end{equation}

Further, the group convolution also has a computational advantage over the standard convolution, which reduces the number of parameters in the network by $\frac{G}{c}$ factor \cite{mamalet2012simplifying}, i.e.,
 \unboldmath\begin{equation}
   \frac{n . n . c.G. H.W}{n . n . c.c .H.W} = \frac{G}{c}  \; \; , \text{where}  \;\; G \leq c
\end{equation} 


The output of the diverse feature extraction phase is then passed to the adaptive feature selection phase.  
\subsection{Adaptive feature selection}
{\color{green!55!blue} }
As images from different classes contains different types of features, CNN also has different feature-map preferences for different images \cite{krizhevsky2009learning}. Hence, giving more focus to the informative feature-maps helps to correctly classify the given image. To accomplish that, we evaluate the extracted feature-maps and generate weighting values to adaptively evaluate the feature maps based on their contributions. To achieve this goal, we use global average pooling  to aggregate the global contextual information of $Z^{\mathcal{D}}_{\ell} \in \mathbb{R}^{H\times W\times C}$ along the height and width axis and embed the information in a compact vector representation $ Z^{ \hat{\mathcal{D}}}_{\ell} \in \mathbb{R}^{1,1,c}$, which can be written as:
\begin{equation}
    Z^{ \hat{\mathcal{D}}}_{\ell} =  \frac{1}{H \times W} \sum \limits_{i=1}^{H} \sum \limits_{j=1}^{W}  Z^{\mathcal{D}}_{\ell} (i,j).
\label{eq:maping}
\end{equation}
\unboldmath

Since each feature-map holds a distinctive feature detector \cite{zeiler2014visualizing}, we aim to focus on learning the inter-feature-map relationship at the local level among adjacent feature-maps. Let $\omega$ be the parameter we want to learn for adaptive selection.
We compute each $\omega$ by considering the information among $\alpha$ neighboring feature-maps in $Z^{\hat{\mathcal{D}}}_{\ell}$ to learn the cross-feature-map relationship and generate weighting values $ Z^{\mathcal{A}}_{\ell} \in \mathbb{R}^{1,1,c}$, which can be expressed as, 
\begin{equation}
    Z^{\mathcal{A}}_{\ell} = \sigma \left( \sum \limits_{i=1}^{\alpha} \omega^{i} Z^{\hat{\mathcal{D}}}_{\ell}(i) \right)  \;\; \text{where}   \;\; Z^{\hat{\mathcal{D}}}_{\ell}(i) \in \psi^{\alpha}
    \label{eq:chan_atten}
\end{equation}
where $\psi^{\alpha}$ denotes the set of $\alpha$ neighboring feature-maps in $Z^{\hat{\mathcal{D}}}_{\ell}$, $\sigma$ denotes a sigmoid activation function.

We consider $\alpha$ in range of $\alpha=2y+1$ and $y$ is an integer value ranging in $1\leq y \leq 7$. We limit $y \leq 7$, based on the conducted empirical analysis, which is discussed in Section \ref{sec:experiment}. This strategy can be empirically implemented using a $1D$ convolution with size of $\alpha$, which can be represented as:
\begin{equation}
      Z^{\mathcal{A}}_{\ell} = \sigma( f^{\alpha} ( Z^{ \hat{\mathcal{D}}}_{\ell} ) )
    \label{eq:chan_atten}
\end{equation}
where $f^{\alpha}$ denotes a $1D$ convolution with size of $\alpha$. The output of the adaptive selection phase is passed through a sigmoid function such that the weighting values are squeezed between $0$ and $1$.

Then, the output of adaptive selection $Z^{\mathcal{A}}_{\ell}$ scales the corresponding feature-map from the diverse feature extraction phase $ Z^{ \mathcal{D}}_{\ell}$ via a point-wise multiplication $\circ$. This can be represented as:
\begin{equation}
 Z_{\ell} = Z^{\mathcal{A}}_{\ell}  \circ  Z^{ \mathcal{D}}_{\ell}
    \label{eq:concatination}
\end{equation}
The intuition behind Eq. \ref{eq:concatination} is when the values in $  Z^{\mathcal{A}}_{\ell}$ are close to zero the corresponding feature-map will approach to a very small number, consequently, the feature map becomes off. On the other hand, when $  Z^{\mathcal{A}}_{\ell}$  approaches to $1$ the corresponding feature-map in $Z^{ \mathcal{D}}_{\ell}$ will approximate to its original value. 

In summary, diverse feature extraction and adaptive selection mechanism jointly learn to capture diverse features and understand the cross-feature-map relationship to effectively select the most informative feature-maps and efficiently boost CNN's performance. We conducted extensive experimental analysis and presented the results in the next section. 

\section{Experiments} \label{sec:experiment}
\unboldmath
\begin{figure}[!t]
    \centering
    \includegraphics[width=.5\textwidth]{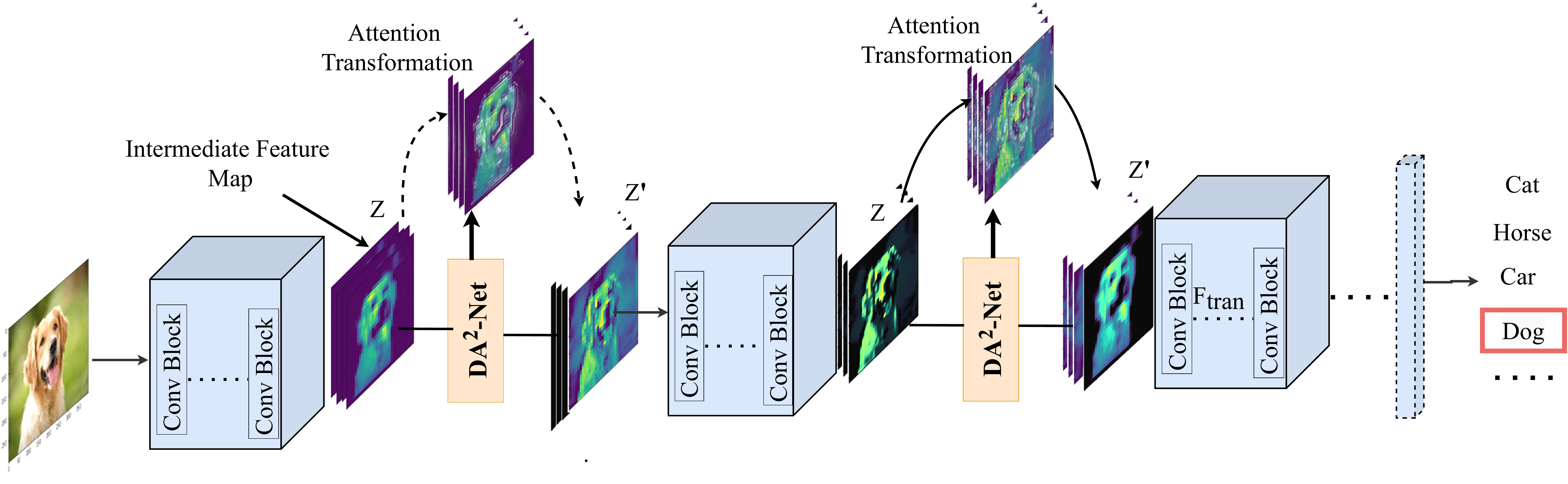}
    \caption{DA$^{2}$-Net integrates with a CNN architecture at the bottleneck location.}
    \label{fig: DANet integration}
\end{figure}
\subsection{Datasets}
The proposed method is evaluated on the following benchmark datasets: 

\textbf{CIFAR-100 \cite{krizhevsky2009learning}}
The CIFAR-100 dataset contains $60$k colored images with $32\times 32$ image size and $100$ number of classes, where $50$k of images are for training and $10$k for testing. We follow the standard data pre-processing as in \cite{he2016deep}. 

\textbf{SVHN \cite{netzer2011reading}}: SVHN dataset contains $630,420$ colored images with $32\times 32$ size and $10$ classes. A standard pre-processing is used as in \cite{goodfellow2013maxout}.

\textbf{ImageNet \cite{krizhevsky2012imagenet}}: The ImageNet dataset contains $1.28$ million training and $50$k validation images with $1,000$ classes. A standard data augmentation is used with random size cropping of $224\times 224$ and with random horizontal flip \cite{szegedy2015going}. A standard pre-processing is used, where the data is normalized using RGB mean and standard deviation values.

\subsection{Experimental Setup}
\unboldmath
Our experimental analysis consists of two parts. We first conducted an ablation study, then we integrated the final DA$^{2}$-Net design with various CNN architectures and compared its effectiveness on benchmark datasets. The comparison is made based on different evaluation metrics, including accuracy, floating-point operation (FLOPs), the number of parameters, inference time, and throughput.

To provide fair comparisons, we first reproduced the reported performance of the CNN baseline architectures \cite{he2016deep,huang2017densely,zagoruyko2016wide,hu2018squeeze} in the PyTorch framework \footnote{https://pytorch.org/} and set the baselines. Note that every model is trained from scratch without using pre-trained network parameters. 
We followed a standard training procedure, where we used the weight initialization method from study \cite{he2015delving} and Stochastic Gradient Decent (SGD) optimizer. We used mini-batch size of $128$ for CIFAR-100 and SVHN and $256$ for ImageNet, respectively. The weight decay set to $0.0001$ and the momentum to $0.9$. We set the initial learning rate is to $0.1$ and it decreased by a factor of $10$ per $60$ epochs for CIFAR-100 and SVHN and per $30$ epochs for ImageNet. The models are trained for $180$ epochs on CIFAR-100 and SVHN and for $90$ epochs on ImageNet. We used two NVIDIA Tesla V$100$ and two NVIDIA P$100$ GPUs to generate result for the ImageNet dataset, and an NVIDIA Tesla P$100$ GPU to generate result for the SVHN and CIFAR100 dataset. 

\subsection{Ablation studies on CIFAR-100 dataset}
In this part, we investigated each design component of DA$^{2}$-Net based on widely used CNN architecture ResNet-50 \cite{he2016deep} and CIFAR-100 dataset\cite{krizhevsky2012imagenet}. Since the classes in CIFAR-100 are highly overlapped, we choose it for the ablation study. As shown in Figure \ref{fig: DANet integration}, we integrated DA$^{2}$-Net at the bottleneck location of CNN architectures. Bottleneck refers as the locations in the architecture where the height and width of the feature maps reduce. For instance, Resnet-50 \cite{ren2015faster} architecture has four bottleneck locations, so we integrate the attention module at those bottlenecks. Similar approach has been used in \cite{park2018bam,woo2018cbam}. 

\textbf{Effect of Filter Size.} 
As shown in Eq. \ref{eq:spatial_atten}, filter size is one of the crucial elements in DA$^{2}$-Net's spatial attention transformation. To study its effects, we compare different filter combinations in the range of $3\leq n \leq 9$,where $n$ is an integer value, with grouping ratio of $g=1$. The filters in DA$^{2}$-Net's layers are placed in an ascending order based on their size. For instance, if we use filters with size of $3,5,7$, their placement would be $3\xrightarrow{} 5 \xrightarrow{} 7$. This results in five unique filter combinations, where four of them are based on three filters and one is based on the combination of four filters.  The output features from the filters then adaptively selected by the feature-map attention, as given in Eq. \ref{eq:chan_atten} and \ref{eq:spatial_atten}. The result are presented in Table \ref{Table:two_filter_comnb1}.

\begin{table}[!h]
\small
\unboldmath
\caption{Results of DA$^{2}$-Net with different combinations of multiple filters; and with ($\checkmark$) and without($\times$) Adaptive selection. Baseline$+$ represents the baseline model (ResNet-$50$) with additional convolutional blocks with the same design as in the baseline model.}
\begin{tabular}{l|l|l|l|l}
\Xhline{2pt}
 Conv filters & Adaptive & Params(M) &GFLOPs &  Acc\% \\ \hline
Baseline& $\times$  &   $23.71$ & $1.30$  & $73.57$  \\ \cline{1-1}
     Baseline$+$ & $\times$ &  $29.63$  &  $1.59$   & $74.46$    \\ \cline{1-1}  
      \Xhline{.5pt}
  $7$ $\xrightarrow{}$ $7$ $\xrightarrow{}$ $7$& \checkmark & $24.03$  & $1.35$   & $75.67 $     \\ \cline{1-1}
  $9$ $\xrightarrow{}$ $9$ $\xrightarrow{}$ $9$& \checkmark  & $24.67$  & $1.43$   & $75.76$ \\ \cline{1-1}
     $3$ $\xrightarrow{}$ $5$ $\xrightarrow{}$ $7$&  \checkmark & $\textbf{24.05}$    & $\textbf{1.35}$ & $\textbf{77.86}$   \\ \cline{1-1}
   $5$ $\xrightarrow{}$ $7$ $\xrightarrow{}$ $9$& \checkmark &  $24.33$    & $ 1.39$ & $76.37$   \\ \cline{1-1}
     $3$ $\xrightarrow{}$ $7$ $\xrightarrow{}$ $9$&  \checkmark & $24.27$   & $1.38$ & $76.77$    \\ \cline{1-1}
    $3$ $\xrightarrow{}$ $5$ $\xrightarrow{}$ $9$&  \checkmark & $24.18$    & $1.37$ & $76.86$  \\ \cline{1-1}
     $3 \tiny{\xrightarrow{}}5 \tiny{\xrightarrow{} }7 \tiny{\xrightarrow{}}9$ & \checkmark  & $24.38$   & $1.40$ & $75.50$  \\ \cline{1-1}   
     \hline
       \Xhline{1pt}
   $3$$\xrightarrow{}$$5$ $\xrightarrow{}$ $7$&  $\times$ & $24.05$    & $1.35$ & $76.16$   \\ \cline{1-1}
   \hline
      \Xhline{.5pt}
   \cline{1-1}
  \Xhline{2pt}
\end{tabular}
\label{Table:two_filter_comnb1}
\end{table}

As shown in Table \ref{Table:two_filter_comnb1}, DA$^{2}$-Net effectively improved the baseline performance. We can observe the following points from Table \ref{Table:two_filter_comnb1}: 
\begin{itemize}
    \item When multiple filter size combinations are used, a significant performance improvement is achieved. However, when a similar filter size is used in each layer of  DA$^{2}$-Net's (e.g $7\xrightarrow{}7\xrightarrow{}7$), the performance drops. This indicates the importance of using multiple-size filter combinations to effectively capture diverse features to boost CNN's performance. The $3\xrightarrow{}5\xrightarrow{}7$ filter combination achieved the best performance improvement; hence we use it for the subsequent ablation study. 

    \item The adaptive selection has a significant impact on accuracy. As shown in Table \ref{Table:two_filter_comnb1} , the accuracy of the selected filter combination $3\xrightarrow{}5\xrightarrow{}7$ drops, when the adaptive selection mechanism is removed.  
    \item We empirically verify that the performance gain achieved by DA$^{2}$-Net is not attributed to the increased number of parameters; instead, it is from the DA$^{2}$-Net design principles as discussed in Section \ref{sec:methodolody}. To demonstrate that, in-place of DA$^{2}$-Net, we inserted convolutional blocks with the same design as in the baseline architecture, referred to as Baseline$+$ in Table \ref{Table:two_filter_comnb2}. We can observe that DA$^{2}$-Net significantly improves the performance and adds less overhead than naively adding extra convolutional block layers with the same structure as in the baseline. 
\end{itemize}

\begin{table}[!h]
\small
\unboldmath
\caption{Comparison for demensionality reduction used in DA$^{2}$-Net with  $3\xrightarrow{}5\xrightarrow{}7$ filter combination, $g$ denotes grouping ratio.}
\begin{tabular}{l|l|l|l}
\Xhline{2pt}
Dim Reduction& Params(M) &GFLOPs &  Acc\%  \\ \hline
  \Xhline{.5pt}
  
    $g$=$16$  & $28.83$    & $1.96$ & ${77.79}$ \\
    \cline{1-1}
    
      $g$=$8$  & $26.28$    & $1.64$ & $77.71$   \\ \cline{1-1}

    $g$=$4$  & $25.02$    & $1.47$ & $77.63$  \\ \cline{1-1}
    
           $g$=$2$ & $24.37$    & $1.39$ & $77.70$  \\ \cline{1-1}
    
   $g$=$1$  & $\textbf{24.05}$    & $\textbf{1.35}$ & $\textbf{77.86}$   \\ \cline{1-1}
     \hline
       \Xhline{1pt}
    
    Point-Wise Conv  & $24.26$    & $1.38$ & ${76.13}$ \\
    
    \cline{1-1}
        \Xhline{.5pt}
      \Xhline{.5pt}

   \cline{1-1}
  \Xhline{2pt}
\end{tabular}
\label{Table:two_filter_comnb2}
\end{table}

\textbf{Effect of Spatial Attention Group size}. In this part, we investigate the effect of the spatial attention group size $G$ on DA$^{2}$-Net's performance gain. The grouping ratio $g$ is used to control the group size, as $G=c/g$, that minimize DA$^{2}$-Net's computational overhead. In many CNN architectures \cite{howard2017mobilenets,he2016deep,ren2015faster,huang2017densely,simonyan2014very,szegedy2015going,zagoruyko2016wide}, the minimum number of feature-maps mostly starts from $16$ ($c \geq 16$), and increase in a power-of-two. Accordingly, we considered the values of $g \in [1,2,4,8,16]$ to guarantee zero reminder division with the number of feature-maps $c$. A reduction ration of $16$ is used for the point-wise convolution based dimensionality reduction as stated in \cite{park2018bam,woo2018cbam}. 
    
As shown in Table \ref{Table:two_filter_comnb2},  $g=1$ gives the best performance improvement. We can also observe that despite the increase in the number of parameters, increasing $g$ value does not increase the accuracy. On the other hand, the pointwise convolution-based dimensionality reduction, which is used in other attention mechanisms \cite{park2018bam, woo2018cbam}, gives poor performance. This can be attributed to the effect of pointwise convolution in destroying the linear relationship among feature-maps. Hence, we found depthwise-separable convolution ($g=1$) to be the most effective and efficient approach for DA$^{2}$-Net.
  
\begin{figure}[!h]
    \centering
    \includegraphics[width=0.4\textwidth]{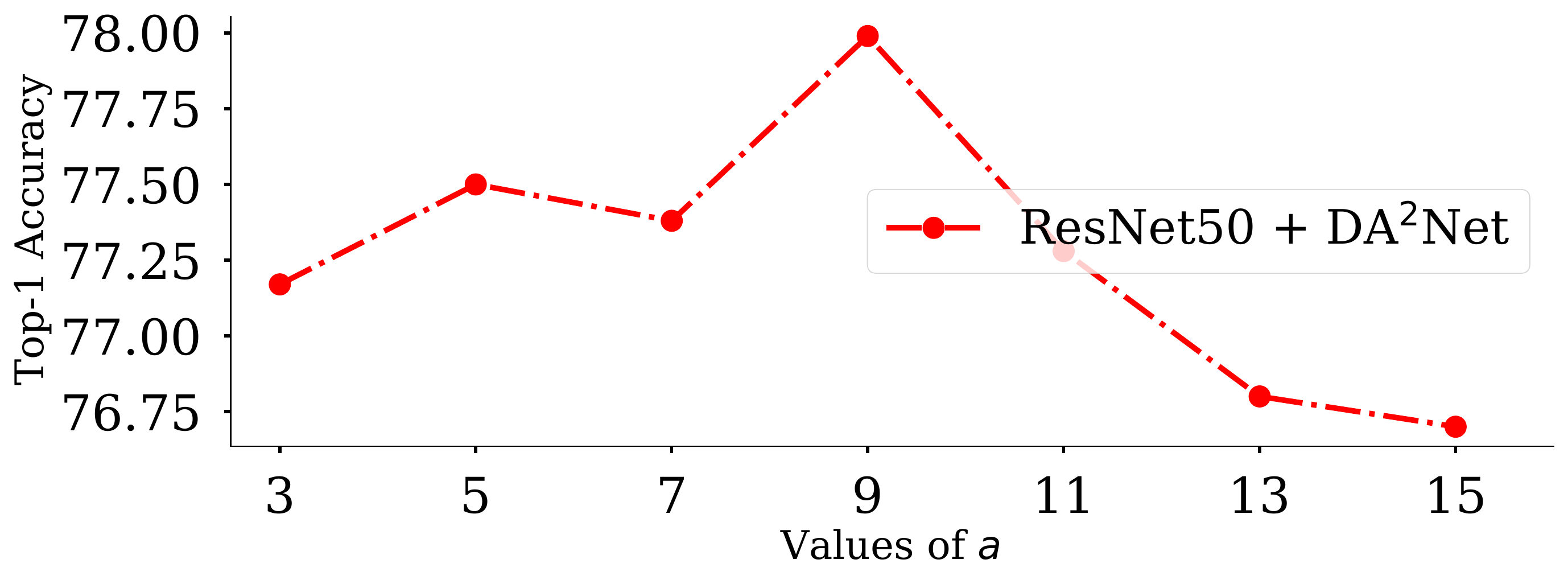}
    \caption{Results of our DA$^{2}$-Net module with various values of $\alpha$ using ResNet-50 as backbone model.}
    \label{fig:DANet_g}
\end{figure}

\textbf{Effect of feature-map Attention Group Size} 
As shown in Equation \ref{eq:chan_atten}, a parameter $\alpha$ determines the number of adjacent local feature-maps DA$^{2}$-Net's feature-map attention needs to consider for learning the cross-feature-map interaction.
We investigated $\alpha$ values starting from $\alpha=3$ and presented the result in Fig. \ref{fig:DANet_g}. From the results, we can observe that as $\alpha$ increases, the performance also increases until $\alpha=9$, then it starts to drop. This indicates that learning local interaction among $9$ feature-maps better enables the DA$^{2}$-Net to capture the local interaction and improves performance effectively.

\subsection{Comparison with Different CNN models and other Attention Mechanisms}

To further evaluate the effectiveness of DA$^{2}$-Net,  we integrated it with state-of-the-art CNN models \cite{he2016deep, zagoruyko2016wide, xie2017aggregated, li2019selective} and conducted performance and efficiency comparison. Also, we compared DA$^{2}$-Net with other similar latest attention mechanisms \cite{park2018bam, woo2018cbam}.

\begin{figure}[th]
    \centering
    \includegraphics[width=0.5\textwidth]{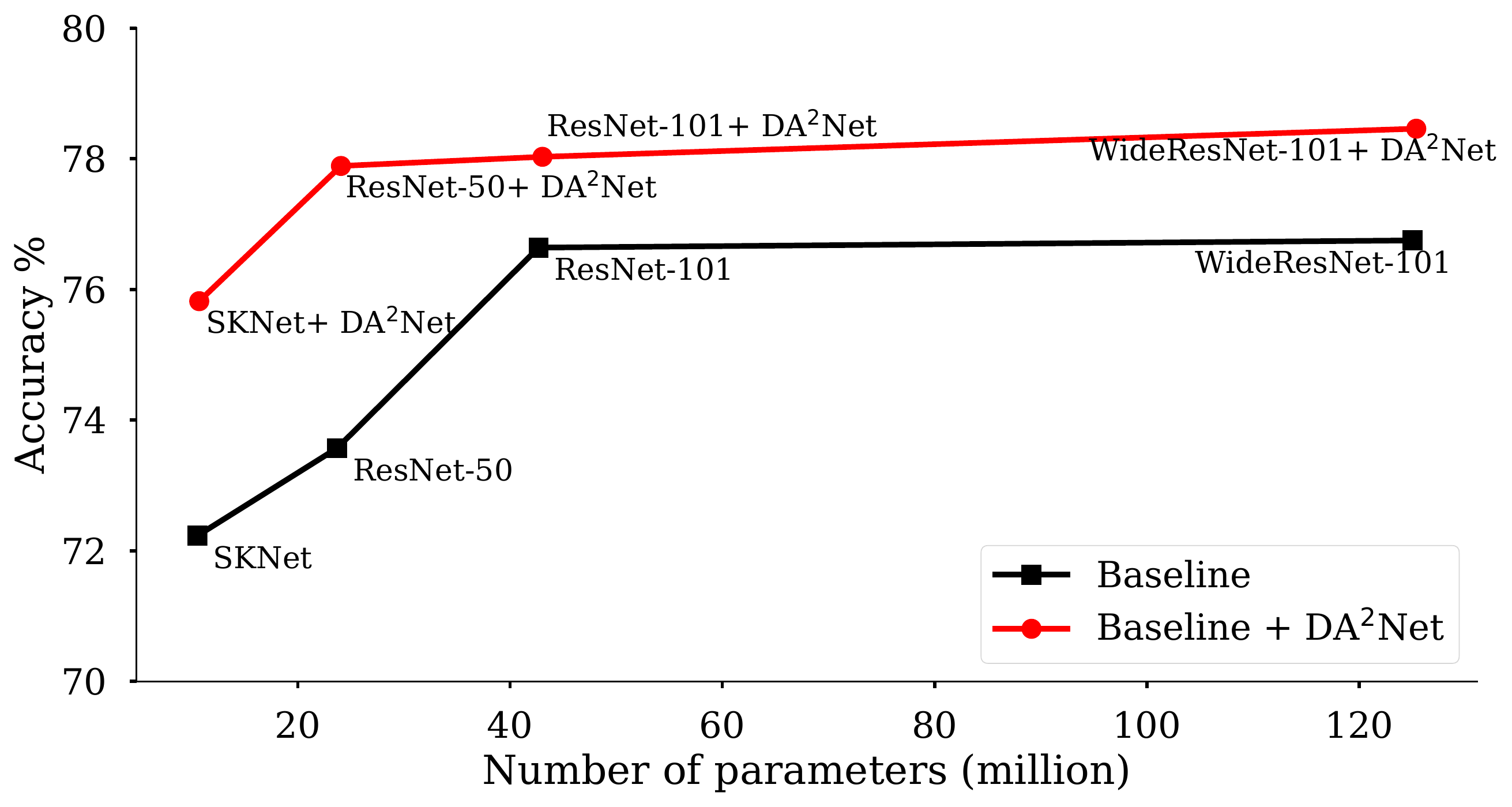}
    \caption{Comparison of baseline models with DA$^{2}$-Net's on CIFAR-100 dataset.}
    \label{fig:acc_comarsion1}
\end{figure}

\begin{table}[th] 
\small
\unboldmath
\caption{Comparison with various models on CIFAR-100 datasets with top-1 validation accuracy.}
\begin{tabular}{l|l|l|l}
\Xhline{2pt}
Architecture & Parm(M) & GFLOPs & Acc\%    \\ 
 \hline \hline
        SKNet \cite{li2019selective} &  $10.54$ & $0.36$ &  $72.23$  \\
            SKNet  \cite{li2019selective}+{\scriptsize SE \cite{hu2018squeeze}} &  $10.91$ & $0.37$  &  $72.19$ \\
    SKNet  \cite{li2019selective}+{\scriptsize BAM \cite{park2018bam}} &  $10.71$ & $0.37$ &  $72.96$  \\
    
    SKNet  \cite{li2019selective}+{\scriptsize CBAM \cite{woo2018cbam}} &  $10.71$ & $0.37$ &  $75.01$  \\

    SKNet  \cite{li2019selective}+\textbf{{\scriptsize DA$^{2}$-Net}}  &  $\textbf{10.71}$ & $\textbf{0.37}$ &  $\textbf{75.82}$ \\
    \hline
       
     ResNet50 \cite{he2016deep} &  $23.70$ & $1.30$ &  $73.57$    \\  
          ResNet50 \cite{he2016deep}+{\scriptsize SE \cite{hu2018squeeze}} & $24.40$  & $1.31$  & $77.28$  \\
    
    ResNet50 \cite{he2016deep}+{\scriptsize BAM \cite{park2018bam}} &  $25.14$ & $1.34$ &  $76.03$  \\
    
   ResNet50 \cite{he2016deep}+{\scriptsize CBAM \cite{woo2018cbam}} &  $24.41$ & $1.31$ &  $76.50$  \\

    ResNet50 \cite{he2016deep}+\textbf{{\scriptsize DA$^{2}$-Net}}  &  $\textbf{24.06}$ & $\textbf{1.35}$ &  $\textbf{77.89}$   \\ \hline 
    
 
    ResNet101 \cite{he2016deep}  &  $42.69 $ & $2.52$ &  $76.64$  \\
    
     ResNet101 \cite{he2016deep}+{\scriptsize SE \cite{hu2018squeeze}} & $43.39$  & $2.52$ & $77.17$  \\
    
    ResNet101 \cite{he2016deep}+{\scriptsize BAM \cite{park2018bam}} &  $44.14$ & $2.56$ &  $73.67$   \\
    
     ResNet101 \cite{he2016deep}+{\scriptsize CBAM \cite{woo2018cbam}} &  $43.39$ & $2.52$ &  $76.63$   \\

    ResNet101 \cite{he2016deep}+\textbf{{\scriptsize DA$^{2}$-Net}}  &  $\textbf{43.05}$ & $\textbf{2.57}$ &  $\textbf{78.03}$   \\    \hline
         
       ResNetXt-$50$ \cite{xie2017aggregated} &  $23.17$ & $ 1.35$ &  $75.73$  \\

      ResNetXt-$50$  \cite{xie2017aggregated}+{\scriptsize SE \cite{hu2018squeeze}} & $23.87$  & $1.35$ &  $78.03$ \\
    
     ResNetXt-$50$  \cite{xie2017aggregated}+{\scriptsize BAM \cite{park2018bam}} &  $24.62$ & $1.39$ &  $76.35$  \\
     
      ResNetXt-$50$  \cite{xie2017aggregated}+{\scriptsize CBAM \cite{woo2018cbam}} &  $23.88$ & $1.35$ &  $78.16$  \\

     ResNetXt-$50$ \cite{xie2017aggregated}+\textbf{{\scriptsize DA$^{2}$-Net}}  &  $\textbf{23.53}$ & $\textbf{1.40}$ &  $\textbf{78.30}$    \\  \hline

    
    
     
      WideResNet-$50$ \cite{zagoruyko2016wide} &  $67.03$ & $3.69$ &  $73.62$  \\
    
         WideResNet-$50$ \cite{zagoruyko2016wide}+{\scriptsize SE \cite{hu2018squeeze}} & $67.72$  & $3.69$  &  $78.12$ \\
    
     WideResNet-$50$ \cite{zagoruyko2016wide}+{\scriptsize BAM \cite{park2018bam}} &  $68.47$ & $3.73$ &  $73.72$  \\
     
     WideResNet-$50$ \cite{zagoruyko2016wide}+{\scriptsize CBAM \cite{woo2018cbam}} &  $67.73$ & $3.70$ &  $78.41$  \\

     WideResNet-$50$ \cite{zagoruyko2016wide}+\textbf{{\scriptsize DA$^{2}$-Net}} &  $\textbf{67.38}$ & $\textbf{3.74}$ &  $\textbf{78.53}$    \\  \hline
     
     WideResNet-$101$ \cite{zagoruyko2016wide} &  $125.03$ & $7.41$ &  $76.75$  \\
     
  WideResNet-$101$  \cite{zagoruyko2016wide}+{\scriptsize SE \cite{hu2018squeeze}} & $125.73$  & $7.41$ &  $78.04$ \\
          
     WideResNet-$101$  \cite{zagoruyko2016wide}+{\scriptsize BAM \cite{park2018bam}} &  $126.47$ & $7.44$ &  $71.58$  \\
     
      WideResNet-$101$  \cite{zagoruyko2016wide}+{\scriptsize CBAM \cite{woo2018cbam}} &  $125.74$ & $7.41$ &  $77.96$  \\
      
     WideResNet-$101$ \cite{zagoruyko2016wide}+\textbf{{\scriptsize DA$^{2}$-Net }}  &  $ \textbf{125.38}$ & $\textbf{7.45}$ &  $\textbf{78.46}$    \\  \hline
   
             \Xhline{2pt}
      \end{tabular}
\label{tab:classification_Cifar_100}
\end{table}

\begin{figure}[th]
    \centering
    \includegraphics[width=0.4\textwidth]{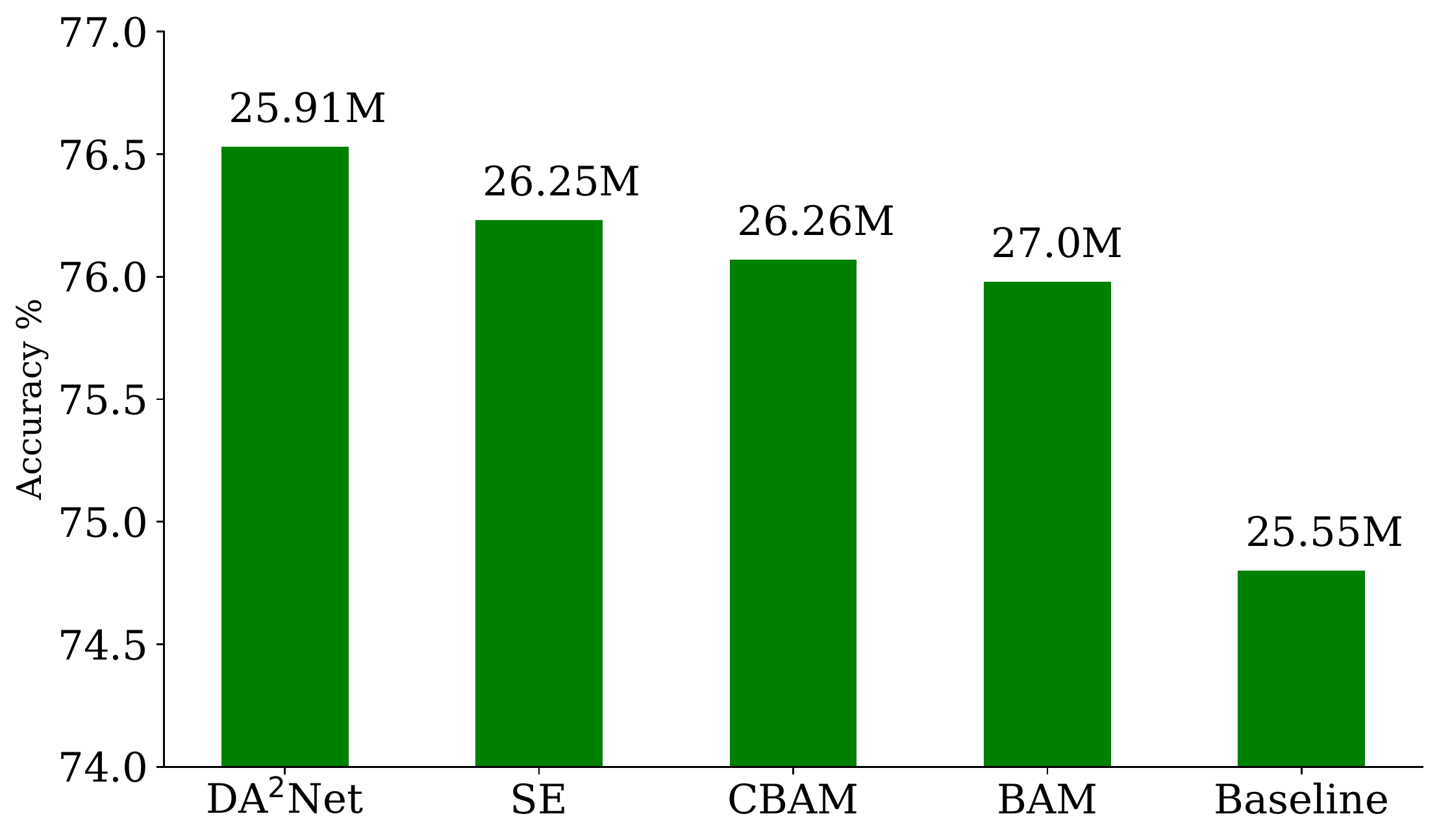}
    \caption{Comparison of top-1 accuracy  on ImageNet \cite{krizhevsky2012imagenet} dataset with ResNet-50 as a baseline model. The model's number of parameter (million) is shown on top of each bar.}
    \label{fig:acc_img}
\end{figure}
\begin{table}[th] 
\small
\unboldmath
\caption{Comparison with various models on SVHN datasets with top-1 validation accuracy.}
\begin{tabular}{l|l|l|l}
\Xhline{2pt}
Architecture & Parm(M) & GFLOPs &  Acc\%   \\ 
 \hline \hline
                  
       AnyNetX \cite{radosavovic2020designing} &  $2.35$ & $0.22$ &  $95.80$     \\
    
     AnyNetX \cite{radosavovic2020designing}  + {\scriptsize BAM \cite{park2018bam}} &  $2.35$ & $0.22$ &  $95.12$  \\
    
      AnyNetX \cite{radosavovic2020designing} + {\scriptsize CBAM \cite{woo2018cbam}} &  $2.41$ & $0.24$ &  $96.13$  \\
    
      AnyNetX \cite{radosavovic2020designing}  + DA$^{2}$-Net &  $\textbf{2.35}$ & $\textbf{0.23}$ &  $\textbf{96.40}$   \\  \hline
      
         ResNet50 \cite{he2016deep} &  $23.52$ & $1.05$ &  $93.21$     \\
    
    ResNet50 \cite{he2016deep} + {\scriptsize BAM \cite{park2018bam}} &  $24.96$ & $1.081$ &  $88.19$  \\
    
     ResNet50 \cite{he2016deep}  + {\scriptsize CBAM \cite{woo2018cbam}} &  $24.22$ & $1.053$ &  $95.09$  \\
    
     ResNet50 \cite{he2016deep}  + DA$^{2}$-Net &  $\textbf{23.87}$ & $\textbf{1.081}$ &  $\textbf{95.72}$   
     
     \\  \hline
     
       ResNetXt-50  \cite{xie2017aggregated} &  $22.99$ & $1.09$ &  $93.54$     \\
    
    ResNetXt-50  \cite{xie2017aggregated} + {\scriptsize BAM \cite{park2018bam}} &  $24.43$ & $1.12$ &  $93.66$  \\
    
     ResNetXt-50  \cite{xie2017aggregated} + {\scriptsize CBAM \cite{woo2018cbam}} &  $23.69$ & $1.088$ &  $95.36$  \\
    
     ResNetXt-50  \cite{xie2017aggregated}  + DA$^{2}$-Net &  $\textbf{23.87}$ & $\textbf{1.081}$ &  $\textbf{96.38}$   \\  \hline

             \Xhline{2pt}
      \end{tabular}
\label{tab:classification_SVHN}
\end{table}

As presented in Table \ref{tab:classification_Cifar_100}, \ref{tab:classification_SVHN} and Fig  \ref{fig:acc_comarsion1},\ref{fig:acc_img}, DA$^{2}$-Net evaluated on three different datasets and achieves a consistent performance improvement with very negligible overhead. For instance, as shown in  Table \ref{tab:classification_Cifar_100}, by integrating DA$^{2}$-Net with ResNet-$50$, a $4.32\%$ of accuracy gain can be achieved, with only a $1.52\%$ increase in the number of parameters and $3.84\%$ increase in GFLOPs. Whereas increasing the depth of ResNet\cite{he2016deep} from $50$ layers (ResNet-50) to $101$ layers (ResNet-$101$) achives a $2.16\%$ performance gain, but with significant increase in computational cost; $80.12\%$ increase in the number of parameters and $93.52\%$ increase in GFLOPs (doubled the computational cost). This clearly shows how DA$^{2}$-Net can better boost CNN's performance with a fraction of the cost. In general, as shown in the presented results DA$^{2}$-Net outperforms baseline models and other attention mechanisms with better and consistent performance improvement. 

\begin{figure}[th]
    \centering
    \includegraphics[width=0.45\textwidth]{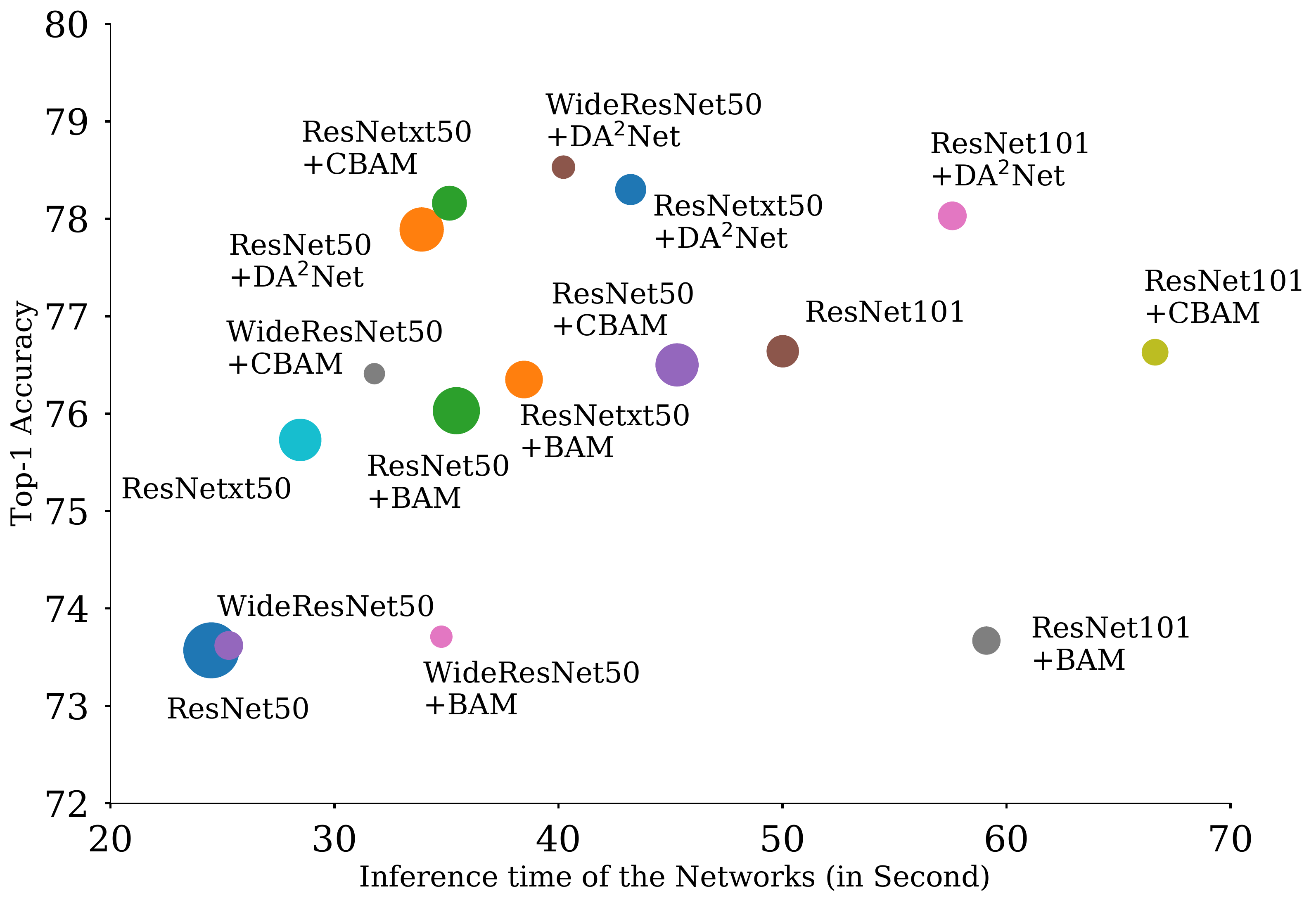}
    \caption{DA$^{2}$-Net's accuracy, inference time, and throughput with other attention mechanisms and baseline models. The size of the circles represents the throughput of the model, where the wider the circle indicates the higher throughput.}
    \label{fig:inff}
\end{figure}
\subsection{Statistical Analysis}
\begin{figure}[!h]
    \centering
    \includegraphics[width=0.45\textwidth]{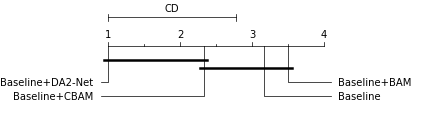}
    \caption{CD diagram for Nemenyi test with  0.05 significance level.}
    \label{fig:stat}
\end{figure}
\textbf{Inference time and Throughput.} 
In this experiment, we systematically examined and compared the inference time and throughput of DA$^{2}$-Net with other attention mechanisms and baseline models. For each experiment to avoid the GPU running into the power-saving mode, the GPU and the CPU are synchronized then warmed up by running dummy examples. We calculated the throughput ($\tau$) as: $\tau = \frac{N . b}{t} $, where N, b, and t denotes the number of batches, the maximum batch size, and the total recorded time in seconds, respectively. We run each experiment for $100$ times, and we take the average value. As illustrated in Fig. \ref{fig:inff}, compared with other attention mechanisms\cite{park2018bam,woo2018cbam}, DA$^{2}$-Net results in better accuracy improvements (up-to $+2.5\%$ increase), with less or comparable increase in inference time ($\pm 1.5\%$) and decrease in throughput ($\pm 2.8\%$). 

\textbf{Friedman rank test.} To conduct the Friedman rank test \cite{demvsar2006statistical}, we rank each model accuracy over different architectures and compute their average ranks. From the Friedman test, we get a p-value of $0.00384 < 0.05$, indicating that all compared methods have statistically different performance. As shown in Fig. \ref{fig:stat}, the Nemenyi post-hoc test \cite{demvsar2006statistical} reveals that the proposed DA$^2$-Net method shows statistically significant performance improvement than BAM and baseline models and statistically comparable performance with CBAM. The critical distance (CD) serves as the threshold to determine whether there is a statistically significant difference between each pair of compare methods. In Fig \ref{fig:stat}, methods with statistically comparable performance are connected with the solid line while methods with a significant difference are disconnected.

\section{Conclusion and Future Work} \label{sec:conclusion}
This paper proposed DA$^{2}$-Net to efficiently boost Convolutional Neural Networks (CNNs) performance by capturing diverse features and adaptively selecting and emphasizing the most informative features. DA$^{2}$-Net is designed to be easily integrated with any CNN architecture. We conducted extensive experiments with various CNN architectures based on a well-known benchmark dataset, including CIFAR-100, SVHN, and ImageNet-1K. Experimental results show DA$^{2}$-Net consistently improves CNN's performance with negligible computational cost and outperforms other attention mechanisms. In future work, we will implement DA$^{2}$-Net for object detection and segmentation task. We will further investigate the effectiveness of DA$^{2}$-Net for challenging aerial perception task, including for small drone detection and classification application.

\section*{ACKNOWLEDGMENT}
This work is supported by the National Institute of Aerospace's Langley Distinguished Professor Program under grant number C16-2B00-NCAT. This work is also partially funded by the NASA University Leadership Initiative (ULI) under grant number $80NSSC20M0161$, Air Force Research Laboratory and the OSD under agreement number FA8750-15-2-0116, OSD RTL under grants number W911NF-20-2-0261.
\bibliographystyle{IEEEtran}
\bibliography{smc}

\end{document}